\documentclass[preprint]{article}

\usepackage{neurips_2025}

\usepackage[utf8]{inputenc} 
\usepackage[T1]{fontenc}    
\usepackage{hyperref}       
\usepackage{url}            
\usepackage{booktabs}       
\usepackage{amsfonts}       
\usepackage{nicefrac}       
\usepackage{microtype}      
\usepackage{xcolor}         
\usepackage{comment}
\usepackage{amsmath}        
\usepackage{graphicx}
\usepackage{makecell}
\usepackage{float}

\usepackage{xcolor}
\newcommand{\up}{\textcolor{gray}{$\uparrow$}}
\newcommand{\down}{\textcolor{gray}{$\downarrow$}}
\hypersetup{
hidelinks,
colorlinks=true,
linkcolor=black,
citecolor=black
}

\title{Spark-Prover-X1: Formal Theorem Proving Through Diverse Data Training}

\author{%
  Xinyuan Zhou~~~~~~Yi Lei~~~~~~Xiaoyu Zhou~~~~~~Jingyi Sun\thanks{~Work done during internship at iFlytek Research.}~~~~~~Yu Zhu\footnotemark[1]\\
  ~\bf Zhongyi Ye~~~~~~Weitai Zhang~~~~~~Quan Liu\thanks{~Contact person: {quanliu@iflytek.com}.}~~~~~~Si Wei~~~~~~Cong Liu\\
  \\
  ~iFlytek Research \\
}

\begin{document}

\maketitle

\begin{abstract}

Large Language Models (LLMs) have shown significant promise in automated theorem proving, yet progress is often constrained by the scarcity of diverse and high-quality formal language data. To address this issue, we introduce Spark-Prover-X1, a 7B parameter model trained via an three-stage framework designed to unlock the reasoning potential of more accessible and moderately-sized LLMs. The first stage infuses deep knowledge through continuous pre-training on a broad mathematical corpus, enhanced by a suite of novel data tasks. Key innovation is a "CoT-augmented state prediction" task to achieve fine-grained reasoning. The second stage employs Supervised Fine-tuning (SFT) within an expert iteration loop to specialize both the Spark-Prover-X1-7B and Spark-Formalizer-X1-7B models. Finally, a targeted round of Group Relative Policy Optimization (GRPO) is applied to sharpen the prover's capabilities on the most challenging problems. To facilitate robust evaluation, particularly on problems from real-world examinations, we also introduce ExamFormal-Bench, a new benchmark dataset of 402 formal problems. Experimental results demonstrate that Spark-Prover achieves state-of-the-art performance among similarly-sized open-source models within the "Whole-Proof Generation" paradigm. It shows exceptional performance on difficult competition benchmarks, notably solving 27 problems on PutnamBench (pass@32) and achieving 24.0\% on CombiBench (pass@32). Our work validates that this diverse training data and progressively refined training pipeline provides an effective path for enhancing the formal reasoning capabilities of lightweight LLMs. We will release both Spark-Prover-X1-7B and Spark-Formalizer-X1-7B, along with the ExamFormal-Bench dataset, in the near future.

\end{abstract}

\section{Introduction}

\begin{figure}
  \centering
  \includegraphics[width=\textwidth]{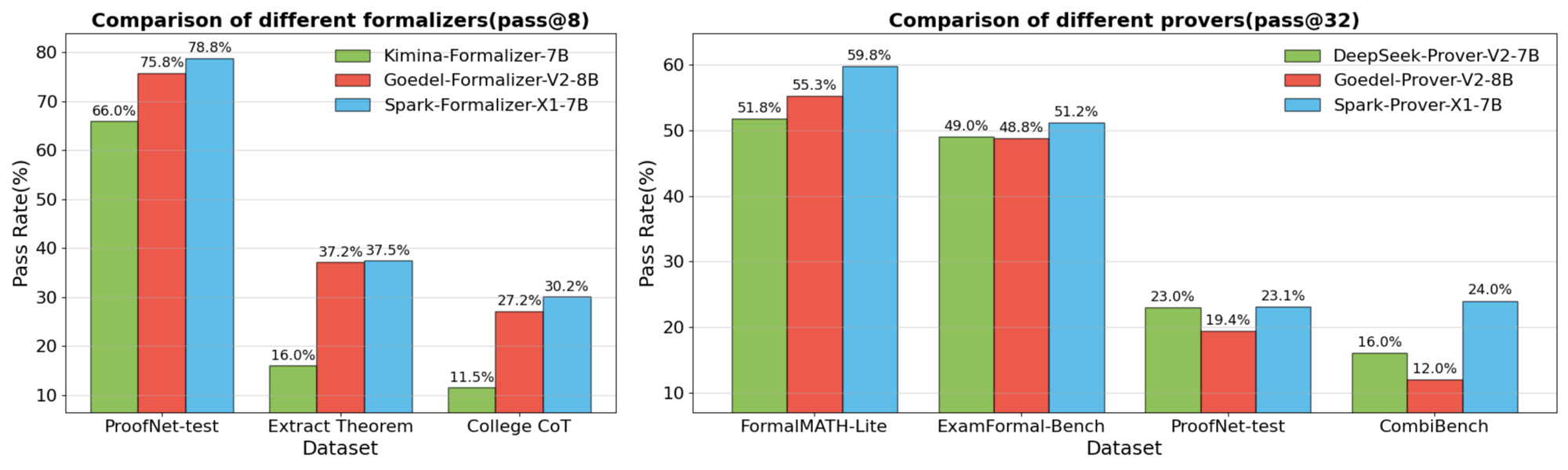}
  \caption{Performance comparison of formalizer and prover models on different datasets.}
  \label{fig:overall_compare}
\end{figure}

Automated Theorem Proving (ATP), as a grand challenge in the field of artificial intelligence, is experiencing new opportunities with the rise of Large Language Models (LLMs)\cite{formalmathematicalreasoningnew}. Formal languages\cite{lean, lean4} can provide clear and reliable supervision signals for LLM training through their rigorous logical frameworks and machine verifiability, making them an ideal testbed for exploring advanced machine reasoning capabilities\cite{gptf}. Formal theorem proving demands that a model not only possesses deep mathematical intuition but also masters precise symbolic language syntax, and can conduct long-chain strategic planning and logical deduction within a vast search space\cite{theoremllama}.

However, the issue of incomplete knowledge coverage and data bias\cite{internlm2} limits the progress of LLM in formal proof generation, due to the scarcity of high-quality formal training data. Acquiring high-quality formal theorem proving data is both expensive and difficult\cite{stp, deepseekproverv1, hunyuanprover}, and significant distributional differences exist between established mathematical libraries and the problems found in common benchmarks\cite{goedelv1}, which limits a model's generalization capabilities. Existing works mainly adopt expert iteration with cold-starting by a small amount of data, escaping the formal data pre-trained phase. For this paradigm, it is almost impossible to sufficiently utilize the various-style and multi-task formal open-source data.

To this end, we introduce Spark-Prover-X1-7B, a large language model with excellent formal theorem proving performance. The Spark-Prover-X1-7B adopts the three-stage training strategy, i.e., pre-training, Supervised Fine-tuning (SFT), and reinforcement learning (RL).
To provide a solid foundation for auto-formalization and automated theorem proving, we inject a broad spectrum of mathematical data into the model through the Continue Pre-training stage\cite{continuelearning}. 
Subsequently, we simultaneously develop a prover and formalizer through Supervised Fine-tuning (SFT) focusing on auto-formalization and theorem proving, respectively. The expert iteration framework\cite{expertiteration} is also employed for both pre-training and SFT stages. After this iterative process, we apply a single round of reinforcement learning (RL)\cite{reinforcement} with GRPO\cite{grpo} to further strengthen the prover's ability to solve complex formal problems.

Our contributions can be primarily summarized in the following:

\begin{itemize}

\item[$\bullet$] \textbf{Enhanced pre-train model.} We improve performance of the pre-train model on auto-formalization and automated theorem proving and simultaneously retain the natural language understanding capabilities by leveraging a vast of diverse formal and informal data towards multiple math- and formal-related tasks. The various training data covers statement auto-formalization, in-formalization, formal proof generation, informal proof generation, tactic prediction, next state prediction, formal code file, translation of the proof, subgoal decomposition, tutorial of tactics/Mathlib, and informal mathematics problem solving.

\item[$\bullet$] \textbf{Novel data augmentation.} We propose a novel data augmentation method to enhance the code only data. The code is modified to Cot-augmented state prediction data, which further explores the step-wise reasoning logic in the proof. This fine-grained data augmentation improves the LLM performance on proof generation.

\item[$\bullet$] \textbf{Open-sourced formal language benchmark.} We introduce a formal language benchmark dataset, named ExamFormal-Bench, consists of 402 problems focusing on formal theorem proving. The ExamFormal-Bench is designed to concentrate on evaluations of problem solving abilities in exams of high school, undergraduate and corresponding competitions.

\end{itemize}

Experimental results validate our framework, demonstrating that both our 7B \texttt{formalizer} and \texttt{prover} establish new state-of-the-art performance among open-source models of a comparable size. Spark-Formalizer-X1-7B achieves the highest average pass rate of 68.57\%, with standout results on \textbf{ProofNet-test} (78.85\%). Concurrently, Spark-Prover-X1-7B achieves a leading 37.0\% average pass rate, demonstrating strong performance on difficult competition benchmarks by solving 27 problems on \textbf{PutnamBench} and achieving 24\% on \textbf{CombiBench}.

\section{Related Work}

\paragraph{LLM-based Automated theorem proving.}

As Large Language Models (LLMs) have demonstrated powerful capabilities in complex reasoning tasks\cite{gpt4techreport}, recent research has widely explored their application in the field of Automated Theorem Proving (ATP). LLM-based automated theorem proving is primarily divided into two technical paradigms. The first is \textbf{Step-wise Tactic Generation}, where the model iteratively generates single proof steps (tactics) during the proving process and interacts with a proof assistant in real-time\cite{internlm25, bfsv2, hunyuanprover, mps}. This paradigm is typically combined with search algorithms such as Best-First Search (BFS)\cite{bfs} or Monte Carlo Tree Search (MCTS)\cite{mtcs} to explore the vast proof space. The second is \textbf{Whole-Proof Generation}, where the model generates a complete proof script in a single pass based on a given theorem statement\cite{baldur, seedprover, goedelv2, deepseekproverv2, kiminapre,kimina_prover_2025}. This approach aims to leverage the powerful long-range coherence and high-level planning capabilities of LLMs. Our work follows the whole-proof generation paradigm to fully leverage the global planning capabilities of LLMs.

Supervised Fine-tuning (SFT) is foundational for building a prover's capabilities, while reinforcement learning algorithms like GRPO\cite{grpo} and preference optimization algorithms such as DPO\cite{dpo} are used to further break through performance bottlenecks\cite{leanabellv1, leanabellv2, deepseekproverv15}. To guide the model's learning effectively, the importance of Curriculum Learning has also been demonstrated. For instance, Goedel-Prover proposed ``scaffolded data synthesis'' to generate tasks of increasing difficulty\cite{goedelv2}, while STP implements an adaptive curriculum through a self-play mechanism\cite{stp}. Meanwhile, to reduce training costs, Delta Prover\cite{delta} proposes training-free agentic frameworks that leverage a general-purpose large model for problem decomposition and iterative proof repair at inference time. Existing works have also attempted to improve the capabilities of lightweight large language models through distillation from larger, more powerful models\cite{kiminapre, deepseekproverv2}.

Scaling up model size or employing complex search strategies can significantly improve proving capabilities, but this also leads to higher computational costs and raises the barrier for reproduction or deployment. Lightweight models are more manageable in terms of cost, but their performance is often less competitive. As existing methods continue to optimize these smaller models, it is evident that this is a valuable direction for exploration.

\paragraph{Auto-formalization.}

Data scarcity is a foundational challenge in the field of formal theorem proving. To address this issue, the research community has explored various data synthesis strategies. Auto-formalization is a mainstream approach that translates massive volumes of natural language mathematical problems into formal statements\cite{autoformalization}; works such as DeepSeek-Prover\cite{deepseekproverv1} and Goedel-Prover\cite{goedelv1} have constructed large-scale datasets at the million level using this method. Meanwhile, the expert iteration paradigm has been widely adopted, which involves iteratively expanding the training data and enhancing the model's capabilities by using new proofs discovered by the model itself\cite{internlm25, bfsv2, goedelv2, kiminapre}.

Our work builds upon these established philosophies through the following steps. First, for the pre-training phase, we construct a large-scale training corpus by combining curated data—including multi-level mathematics problems and open-source datasets—with our own synthetically generated proofs. This stage yields a powerful base model. Subsequently, during the post-training phase, this base model serves as the starting point for our Spark-Formalizer-X1-7B, which we enhance through Supervised Fine-tuning (SFT) to enable it to translate informal problems into high-quality formal statements. Our Spark-Formalizer-X1-7B is then optimized by repeating the pre-training and SFT stages within our expert iteration framework.

\section{Method}

Our methodology is structured into three distinct phases to progressively develop and refine our Spark-Formalizer-X1-7B and Spark-Prover-X1-7B. We begin with a comprehensive data generation and continuous pre-training phase (detailed in Section~\ref{sec:pretraining}) to build a strong foundational model. This is followed by a Supervised Fine-tuning (SFT) phase (Section~\ref{sec:SFT}), where both the formalizer and prover models are simultaneously improved. Our expert iteration framework encompasses both the continuous pre-training and SFT stages; after this iterative process is complete and stable SFT models are obtained, we apply a targeted round of reinforcement learning using Group Relative Policy Optimization (GRPO)\cite{grpo} algorithm (Section~\ref{sec:reinforcement_learning}) to further sharpen the Spark-Prover-X1-7B's capabilities on the most challenging problems.

\subsection{Pre-training}
\label{sec:pretraining}

To enhance the performance of our model in multiple formal math-related downstream tasks, such as auto-formalization and theorem proving, we conduct pre-training for the base model. Approaches involved in pre-training include four key components, i.e., open-source data collection, fine-grained data augmentation, task-oriented data generation, and unsolved problems decomposition.

\subsubsection{Open-source Data Collection}
\label{subsec:data-collection}
As shown in the top-left corner of Figure~\ref{fig:pretrain}, we begin by searching and collecting formal data available online as far as possible, aiming to sufficiently explore and utilize the potential of existing open-source formal data. The utilized datasets cover widely used formal languages. We collect formal data towards multiple tasks, including auto-formalization, unformalization, stepwise tactic prediction, whole proof generation, etc., involving diverse formats and contents. Meanwhile, the formal language code file from GitHub repositories and tutorials for Mathlib4 and tactics are also downloaded to improve the formal statement and proof generation ability of our pre-training model. In addition to formal data, mathematical proof problems of natural language are also involved to enhance the logical reasoning ability of our model. These multitasking corpus are altogether to provide the fundamental performance of auto-formalization and theorem proof generation. For each type of data, more than 100 generalized prompts are injected into the corpus for aggregating the formal reasoning with LLMs. 

\subsubsection{Fine-grained Data Augmentation}
\label{subsec:data-augmentation}
Although there has been an increasing emergence of publicly released formal data, most of them only contain a pair of formal statements and corresponding formal proofs, which are sparse in step-wise reasoning information. Each step in the proof is regarded as a ``Tactic'', which can update the state information containing the current known conditions and proving goal in this step, until all goals have been solved. These fine-grained ``state-tactic'' information is beneficial for LLMs to learn the formal reasoning. Therefore, we propose a verifiable data augmentation method based on CoT extension to reinforce the code-only data.

\begin{figure}
  \centering
  \includegraphics[width=\textwidth]{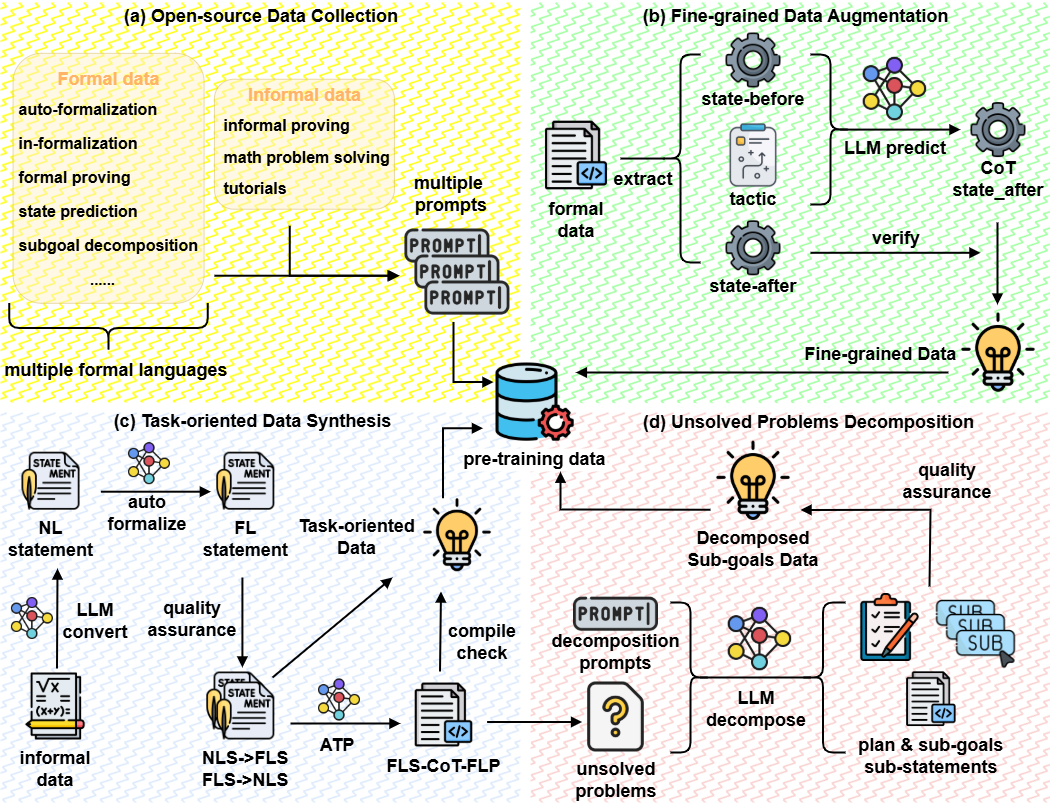}
  \caption{Overview of pretrain process in Spark-Prover-X1-7B. Process (a) is the collection of open-source data using multiple formal language towards multiple tasks. Process (b) is the fine-grained data augmentation by state prediction with CoT. Process (c) is data synthesis orient auto-formalization and formal thoerem proving, where ``NLS'' stands for natural language statement, ``FLS'' means formal language statement, ``NLP'' is natural language proof, and ``FLP'' is formal language proof. Process (d) is the subgoal decomposition with verification for the unsolved problems.}
  \label{fig:pretrain}
\end{figure}

The fine-grained augmented data construction, depicted in the top-right corner of Figure~\ref{fig:pretrain}, includes two processes: 1) tactic-level data extraction and 2) CoT-augmented state prediction. The proposed method decomposes the complete proof code of a proposition into fine-grained tactic-level data and uses LLMs to predict the next state. Then the predicted data is filtered according to the correctness.

\paragraph{Tactic-level Data Extraction.} For a formal dataset $\mathcal{D} = \{ {{Data}^{k}}: k = 1, 2, 3, ..., M\}$ with $M$ pieces of formal data, each data in $\mathcal{D}$ can be denoted as $Data^k = (statement^k, proof^k)$ consisting of theorem statement and proof. We use the formal language proof assistant to extract an information sequence ${info}\_{seq}^k$ containing step-wise tactic and state for $Data^k$. Each element in ${info}\_{seq}^k$ can be organized as a ``state-before, tactic, state-after'' triple:
\begin{equation}
\begin{aligned}
    {info}\_{seq}^k = \{ tp^{k}_{1}, tp^{k}_{2}, tp^{k}_{3}, ..., tp^{k}_{n}\}, 
\end{aligned}
\end{equation}

\begin{equation}
\begin{aligned}
    tp^{k}_{i} = (s\_bf^{k}_{i}, t^{k}_{i}, s\_af^{k}_{i}),
\end{aligned}
\end{equation}
where $n$ is the total number of the proof steps (tactic), $s\_bf^{k}_{i}$ is the state information before the $i$-th tactic $t^{k}_{i}$ conducted in $Data^{k}$, and $s\_af^{k}_{i}$ is the state information after $t^{k}_{i}$ conducted. Obviously, the state after the $i$-th tactic is the same as the state before $(i+1)$-th tactic. For the formal data $Data^k$, concatenating all tactics sequentially can construct the whole formal proof:
\begin{equation}
\begin{aligned}
    proof^{k} = (t^{k}_{1}, t^{k}_{2}, t^{k}_{3}, ..., t^{k}_{n}).
\end{aligned}
\end{equation}
These state information details, feedback from proof assistant, are not visible in the original proof code, which can help the LLM model better understand the formal proof reasoning process.

\paragraph{CoT-augmented state prediction.} 
To further explore the correlation between tactic and state, we introduce the natural language Chain-of-Thought (CoT)-augmented state prediction process.

With the extracted step-wise tactic and state information, we prompt an LLM to predict the next state information with a natural language thinking process. Given states before $s\_bf^k_i$ and current tactic $t^{k}_{i}$, LLM should analyze what will be solved by the tactic. According to the prompt instruction, the LLM then tries to derive what the state will be after applying the tactic on the state before. Both the thinking process $\widehat{CoT}^{k}_{i}$ and final prediction $\widehat{s\_af}^{k}_{i}$ are reserved to construct the CoT-augmented data $\widehat{tp}^{k}_{i} = (s\_bf^{k}_{i}, t^{k}_{i}, \widehat{CoT}^{k}_{i}, \widehat{s\_af}^{k}_{i})$.

To verify the correctness of the predicted data, the predicted $\widehat{s\_af}^{k}_{i}$ is compared with the ground truth ${s\_af}^{k}_{i}$. Those inconsistent results are filtered, and the correctly predicted data should be reserved as the fine-grained Cot-augmented data for the state prediction task during pre-training.

\subsubsection{Task-oriented Data Synthesis}
\label{subsec:synthesis}
The goal of our model is auto-formalization, accurately translating math theorems from natural language to formal language, and theorem proving, automatically proving a proposition in formal language. Existing open-source formal language datasets vary in terms of data quality, content, and knowledge distribution, which limits the performance in auto-formalization and theorem proving. Therefore, to strengthen the pre-training model with greater capability in downstream tasks, we construct a high-quality parallel corpus (see bottom-left of Figure~\ref{fig:pretrain}) oriented towards auto-formalization and theorem proving tasks from a large collection of math problems in natural language.

For the auto-formalization task, our approach transforms informal mathematical problems at high-school and undergraduate levels into theorem format (starting with `Prove that ...') with quality assurance to obtain the informal statement. Then we employ formalizers~\cite{goedelv2, kiminapre} to translate the math proving problems into formal statements. Following previous works~\cite{goedelv1, goedelv2}, the generated formal statements are verified by formal language proof assistant for no syntax error and LLMs for faithful and correct math content. The achieved pair of informal and formal statements is processed for bidirectional translation, i.e., formalization and informalization, for data diversity in pre-training.

For the proof generation task, Deepseek-Prover-V2-7B~\cite{deepseekproverv2} is employed to produce proof of the above generated formal statements. Along with the formal language proof, model's answer also contains a detailed thought process, including problem analysis, informal approach, and formal proposition decomposition. Formal statement and the corresponding proof with CoT construct training data-oriented formal proof generation task.

\subsubsection{Unsolved Problems Decomposition}
Through data synthesis, more than 4 million formal language statements have been produced, but only half of the theorems are solved by prover models. The remaining unsolved problems are too challenging to be proved. To take fully use of these challenging theorems, which are critical for significantly promoting model's proving performance, we decompose them into sub-goals which can be regarded as bridges between the original problems and final proofs. The whole process is shown in the bottom-right part of Fig.\ref{fig:pretrain}.

\paragraph{Sub-goal Decomposition.} 
When prover models fail to solve the theorem, we employ an LLM to conduct a detailed proof plan outlining the main proof steps and strategies in natural language, highlighting key ideas, intermediate lemmas, and proof structures that will guide the construction of the final formal proof. Then, the formal theorem statement is to be decomposed into sub-goals based on the plan. The set of generated sub-goals, when taken together (i.e., their logical conjunction), must be sufficient to prove the original main goal. Each sub-statement is started with ``have'' and ended with ``:= by sorry''. The complete decomposed code is achieved by concatenating the code of all sub-goals ending with an extra ``sorry'' as the placeholder for the proof of the final goal. 

\paragraph{Quality Assurance.} We conduct quality check for the decomposed sub-goals in terms of following criteria: 1) decomposition logic and strategy verification, to check whether the overall approach is valid and the decomposition strategy is reasonable; 2) code consistency verification, to check whether the sub-goals faithfully reflect the informal decomposition and fully comply with formatting requirements; and 3) formal syntax verification, to check whether the decomposed code are approved by the proof assistant. 
The verified decomposed sub-goals can simplify the original problem and empower the pre-training model, which increases possibilities to solve the challenging problems in the post-training phase.

Besides above open-source formal data, state-prediction data, generated translation and proving data, and decomposed sub-goals data, we also introduce challenging math problems and proving problems in natural language with detailed and complete reasoning process. After curating these various training data in proper proportions, the model is pre-trained and shows promise in auto-formalization and formal theorem proving tasks.

\subsection{Supervised Fine-tuning}
\label{sec:SFT}

\begin{figure}
  \centering
  \includegraphics[width=\textwidth]{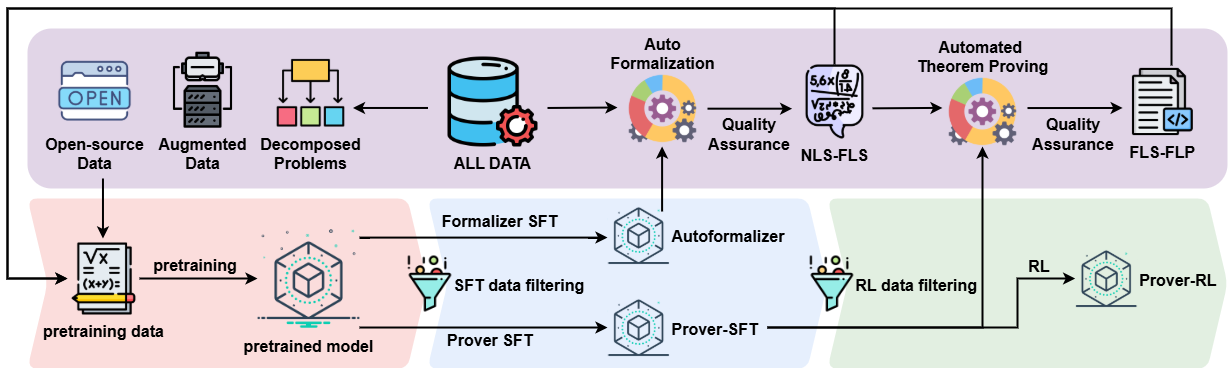}
  \caption{Training framework.}
  \label{fig:framework}
\end{figure}

Based on the pre-training model optimized by various multi-task data, we conduct Supervised Fine-Tuning (SFT) with carefully curated theorem statements for downstream tasks. We employ an expert iteration framework\cite{expertiteration} to simultaneously train and optimize our \texttt{Spark-Formalizer-X1-7B} and \texttt{Spark-Prover-X1-7B} models. As illustrated in Figure~\ref{fig:framework}, our pre-trained model also participates in this iterative loop. The upper half of this framework corresponds to the `Task-oriented Data Synthesis` described in Section~\ref{subsec:synthesis}, which involves using \texttt{formalizer} and \texttt{prover} models for auto-formalization and automated theorem proving. Starting from the second round of the expert iteration, our own fine-tuned models participate in this data synthesis task. This creates a feedback loop that iteratively optimizes the pre-trained model, which in turn serves as an improved base for the next round of SFT, further enhancing the capabilities of our specialized \texttt{Spark-Formalizer-X1-7B} and \texttt{Spark-Prover-X1-7B} models.

\subsubsection{Formalizer}

Starting from the existing large collection of math proving problems in natural language, we conduct dual deduplication. 1) Internal deduplication. For data with cosine similarity exceeding 90\% of the text embeddings, only one piece of data is retained. 2) External deduplication. Deduplication for training data similar to test data is also utilized to achieve reliable results. 

The initial dataset is produced by Goedel-Formalizer-V2-8B~\cite{goedelv2} with reasoning thought for translating the proving statements from natural language to formal language code. The generated formal theorem statements are then verified by proof assistant for syntax check and Qwen3-8B for semantic consistency. The approved data is regarded as training data of SFT in the first-round iteration. In subsequent iteration rounds, our own fine-tuned formalizer is then introduced into the translation pipeline, participating in the expert iteration process to generate new data and further refine its own capabilities.

\subsubsection{Prover}

To achieve excellent ability for formal proof generation, the formal statements available in the pretrain, including open-source and synthesis, are verified to construct the SFT training data. Existing formal statements are filtered according to difficulty and correctness~\cite{goedelv2}. The incorrect and simple theorems are dropped. We also conduct internal and external deduplication on the remaining formal statements.

The formal statements are initialed by DeepSeek-Prover-V2-7B and Goedel-Prover-V2-8B with the proof assistant verification to get the formal proof and corresponding reasoning thoughts. The generated formal proofs must also pass a compilation and a quality inspection. In subsequent iterations, our own fine-tuned Spark-Prover-X1-7B model also participates in the formal proof generation process. It works alongside the other prover models to produce new SFT data, thereby achieving expert iteration optimization.

\subsection{Reinforcement Learning}
\label{sec:reinforcement_learning}

As the final stage of our training methodology, we apply a single round of reinforcement learning (RL) after the expert iteration. We deliberately separate the RL phase from the multi-round Supervised Fine-Tuning (SFT) iterations to allow the model's core capabilities to stabilize and converge before applying a more targeted policy optimization. This strategic separation ensures that the RL process begins with a robust and stable baseline model. This phase also embodies the final step of a curriculum learning strategy that spans our entire training pipeline: from broad knowledge acquisition during unsupervised pre-training, to tackling moderately difficult problems in the SFT-based expert iteration, and finally to focusing on the most challenging theorems in this RL stage. Before initiating the RL training, we first employ a large language model to perform a final difficulty screening on our dataset, selecting only the most difficult subset of problems. This ensures that the computationally intensive RL process is concentrated on the problems that push the boundaries of the model's reasoning capabilities. 

For policy optimization, we adopt GRPO\cite{grpo} as the foundational algorithm and incorporate the clip-higher variant introduced in DAPO\cite{dapo}. To reduce the length bias, we normalize the token-level contribution of each rollout by the maximum sequence length within the group rather than by its own length. We retain the KL regularization term to prevent the updated policy from excessively deviating from the knowledge acquired during SFT. The overall objective is
\begin{equation}
\begin{aligned}
    \mathcal{J}(\theta) &= \mathbb{E}_{q \sim \mathcal{D}, \{o_i\}_{i=1}^G \sim \pi_{\theta_{\text{old}}}} \\
&\frac{1}{G} \sum_{i=1}^G \left\{
\frac{1}{|o|_{max}} \sum_{t=1}^{|o_i|} \left[\min\left(
r_{i,t}(\theta) \hat{A}_i,
\text{clip}\left(r_{i,t}(\theta), 1-\varepsilon_{low}, 1+\varepsilon_{high} \right) \hat{A}_i
\right)
- \beta D_{\text{KL}}
\right] \right\},
\end{aligned}
\end{equation}
where

\begin{equation}
\begin{aligned}
    D_{\text{KL}} = \frac{\pi_{\text{ref}}(o_{i,t} \mid q, o_{i,<t})}{\pi_{\theta}(o_{i,t} \mid q, o_{i,<t})} - \log \frac{\pi_{\text{ref}}(o_{i,t} \mid q, o_{i,<t})}{\pi_{\theta}(o_{i,t} \mid q, o_{i,<t})} - 1,
\end{aligned}
\end{equation}
and

\begin{equation}
\begin{aligned}
    r_{i,t}(\theta) &= \frac{\pi_\theta(o_{i,t} \mid q, o_{i,<t})}{\pi_{\theta_{\text{old}}}(o_{i,t} \mid q, o_{i,<t})}, &\hat{A}_{i}=\frac{R_{i}-\mathrm{mean}(\{R_{i}\}_{i=1}^{G})}{\mathrm{std}(\{R_{i}\}_{i=1}^{G})}.
\end{aligned}
\end{equation}

Here, $|o|_{\max}$ is the maximum sequence length in the group of $G$ sampled trajectories $\{o_i\}_{i=1}^G$. The term $D_{\mathrm{KL}}$ represents a simplified notation for $D_{\mathrm{KL}}(\pi_{\theta}\,\|\,\pi_{\mathrm{ref}})$, and $\pi_{\mathrm{ref}}$ denotes the SFT reference policy. The scalars $\beta$, $\varepsilon_{\mathrm{high}}$, and $\varepsilon_{\mathrm{low}}$ are hyper-parameters. The reward $R_i$ equals 1 if the compiler verifies the rollout’s proof as complete and 0 otherwise. And a soft length penalty\cite{dapo} is additionally applied to the reward. During training, rollouts that exceed the maximum token length are excluded from both advantage computation and subsequent gradient updates. This focused round of policy optimization is instrumental in refining the model's ability to solve the most challenging problems represented in our dataset.

\section{Experiments}

In this section, we present a comprehensive evaluation of our proposed prover and formalizer models. We first detail the full suite of benchmarks used for comparison in Section~\ref{sec:benchmarks}, followed by a description of our training parameter settings in Section~\ref{sec:training_setup}. Finally, we report the main results for our Spark-Formalizer-X1-7B and Spark-Prover-X1-7B in Section~\ref{sec:formalizer_performance} and Section~\ref{sec:prover_performance}, respectively, where we analyze their performance and compare them against existing state-of-the-art, open-source models of a similar scale.

\subsection{Benchmarks}
\label{sec:benchmarks}

To comprehensively evaluate our framework, we conduct experiments on a diverse suite of established and newly introduced benchmarks that span various mathematical domains and difficulty levels. Our evaluation includes widely-used benchmarks such as \textbf{miniF2F}\cite{minf2f}, as well as more specialized and challenging testbeds like \textbf{CombiBench}\cite{combibench} for combinatorial mathematics. Additionally, to better assess model generalization and mitigate potential biases toward specific problem types found in existing benchmarks, we introduce a new test set, \textbf{ExamFormal-Bench}. We release ExamFormal-Bench publicly to contribute a new evaluation resource to the research community. The following provides a detailed description of each benchmark used in our experiments.

\paragraph{miniF2F.} This is a widely recognized benchmark for evaluating formal mathematics systems, which consists of 488 problem statements (244 for validation and 244 for testing), with problems drawn from prestigious competitions such as the American Invitational Mathematics Examination (AIME), the American Mathematics Competitions (AMC), and the International Mathematical Olympiad (IMO), as well as from high-school and undergraduate mathematics courses\cite{minf2f}.

\paragraph{ProofNet.} This benchmark is designed to evaluate performance on undergraduate-level mathematics, comprising 371 formal problem statements in total, with a test set of 186 problems that we use for our evaluation. The problems are primarily sourced from popular undergraduate pure mathematics textbooks and cover topics such as real and complex analysis, linear algebra, abstract algebra, and topology\cite{proofnet}.

\paragraph{PutnamBench.} We evaluate our method on PutnamBench\cite{putnam}, a benchmark comprising 658 competition mathematics problems from the William Lowell Putnam Mathematical Competition. This continuously updated benchmark covers diverse undergraduate domains including algebra, analysis, combinatorics, and number theory, providing a rigorous testbed for automated reasoning systems on challenging mathematical problems.

\paragraph{CombiBench.} CombiBench is a specialized benchmark addressing the scarcity of dedicated benchmarks in combinatorics\cite{combibench}. It comprises 100 combinatorial problems spanning multiple difficulty levels from middle school to IMO and university level, providing a comprehensive testbed for evaluating formal reasoning capabilities in combinatorics.

\paragraph{FormalMATH.} FormalMATH is a comprehensive formal theorem proving benchmark covering diverse mathematical domains including algebra, geometry, calculus, number theory, and discrete mathematics\cite{formalmath}. We evaluate our approach on the FormalMATH-Lite subset, which contains 425 carefully selected problems ranging from high school to undergraduate level.

\paragraph{ProverBench.} ProverBench is a formal theorem proving benchmark comprising 325 problems\cite{deepseekproverv2}. It includes 15 problems from recent AIME competitions focusing on number theory and algebra, 310 problems from curated textbook examples and educational tutorials, providing more comprehensive evaluation across both high-school competition problems and undergraduate-level mathematics.

\paragraph{MathOlympiadBench.} This is a benchmark comprising 360 human-verified formalizations of Olympiad-level mathematical problems\cite{goedelv2}. It aggregates problems from authoritative sources including the IMO (1959-2024), IMO Shortlists, and national mathematical Olympiad problems, providing a comprehensive testbed for evaluating advanced reasoning capabilities.

\paragraph{Extract Theorem \& College CoT.} These two datasets are developed specially for auto-formalization evaluations, each comprising 200 college-level mathematical problems. Extract Theorem comprises theorems sourced from textbooks via OCR covering broad mathematical topics and multilingual content, while College CoT offers LLM-verified problems curated from diverse digital resources\cite{herald}.

\paragraph{ExamFormal-Bench.} We introduce ExamFormal-Bench, which is a formal language benchmark for formal proof generation. It begins with a large-scale question bank, built by collecting problems across middle school, high school, and undergraduate academic levels from exams and competitions, processed via manual OCR. These problems are systematically classified into a structured knowledge taxonomy, including Analysis, Geometry, Algebra, Probability \& Statistics, Computational Mathematics, and Discrete Mathematics. The complete question bank is formed by drawing problems from each of these top-level categories. The final test set contains 402 problems.

\subsection{Training Setup}
\label{sec:training_setup}

Our training process is divided into three stages: continuous pre-training, Supervised Fine-tuning (SFT), and Reinforcement Learning. We first conduct continuous pre-training based on our own mathematical base model, using the Adam optimizer with a learning rate of $4 \times 10^{-5}$ and a maximum token length of 32,768. This initial training is performed on 512 Ascend 910b NPUs to produce the first version of our pre-trained model. In each subsequent round of expert iteration, this entire pre-training process is repeated, starting from the initial mathematical base model but using the continuously updated pre-training dataset, to iteratively optimize the pre-trained model itself.

In the SFT stage, we simultaneously fine-tune our prover and formalizer models. Both models begin from the same pre-trained model checkpoint of the current iteration but are trained on their respective SFT datasets. For each expert iteration round, both the formalizer and prover models are trained for 2 epochs with a global batch size of 128 and a warm-up ratio of 0.002, using the Adam optimizer. The learning rate for both follows a cosine decay schedule, starting from a maximum of $3 \times 10^{-5}$ and decaying to a minimum of $3 \times 10^{-6}$. The primary differences lie in their resource allocation and context length: the Spark-Prover-X1-7B is trained with a maximum token length of 32,768 on 256 Ascend 910b NPUs for approximately 14 hours per iteration, whereas the Spark-Formalizer-X1-7B is trained with a maximum token length of 16,384 on 128 Ascend 910b NPUs for approximately 5 hours per iteration. 

For reinforcement learning, we refine the Spark-Prover-X1-7B with GRPO\cite{grpo} on a curated subset of challenging problems. The RL stage is initialized from the SFT checkpoint and uses a fixed learning rate of $1.5 \times 10^{-6}$ with a fixed KL coefficient of $0.04$. Rewards are provided by a proof assistant adapted from Kimina Lean Server\cite{santos2025kiminaleanservertechnical}, which verifies proof correctness and returns a binary signal. We also apply the soft length penalty from DAPO\cite{dapo} together with its clip-higher variant, setting $\varepsilon_{low} = 0.2$ and $\varepsilon_{high} = 0.28$. Training is conducted with a maximum context length of 32,768 tokens on 320 Ascend 910b NPUs. At each RL iteration, we collect 2,560 queries and generate 16 rollouts per query; optimization uses a global batch size of 320 with gradient accumulation of 4, yielding an effective batch size of 1,280 per optimizer step. Finally, an early stopping strategy is employed to select the best-performing checkpoint.

\subsection{Formalizer Evaluations}
\label{sec:formalizer_performance}

The proposed Spark-Formalizer-X1-7B is evaluated by translating natural language statements into formal language statements on five diverse benchmarks, compared with advanced open-source models on the similar scale. For each natural language proposition, we generate eight translation candidates under pass@8. A formal proposition is considered successfully translated if at least one of the eight candidates passes a rigorous two-stage verification process, successful compilation check and semantic consistency validation using CriticLeanGPT\cite{criticlean}. The detailed results are presented in Table~\ref{tab:autoformalizer-results}.

\begin{table}[H]\footnotesize
  \caption{Comparison results of formalizers under pass@8.}
  \label{tab:autoformalizer-results}
  \centering
  \begin{tabular}{lcccccc}
    \toprule
    Benchmark   & Herald & \makecell{Kimina-\\Formalizer-7B} &  \makecell{Goedel-\\Formalizer-V2-8B} & \makecell{Spark-\\Formalizer-7B} \\
    \midrule
    miniF2F-test & 74.32\% $\pm$ 2.47\%  & 96.58\% $\pm$ 0.63\% &  \textbf{99.18\% $\pm$ 0.82\%} & 98.63\% $\pm$ 0.47\% \\
    ProofNet-test & 54.30\% $\pm$ 1.61\%  & 65.95\% $\pm$ 1.12\% &  75.81\% $\pm$ 0.93\% & \textbf{78.85\% $\pm$ 0.62\%} \\
    Extract Theorem & 8.33\% $\pm$ 0.76\% & 16.00\% $\pm$ 0.87\% &  37.17\% $\pm$ 0.76\% & \textbf{37.50\% $\pm$ 2.00\%} \\
    College CoT & 6.83\% $\pm$ 0.76\%  & 11.50\% $\pm$ 0.50\% & 27.17\% $\pm$ 0.58\% & \textbf{30.17\% $\pm$ 1.26\%} \\
    ExamFormal-Bench & 67.91\% $\pm$ 1.51\% & 85.32\% $\pm$ 1.32\% & 97.43\% $\pm$ 0.52\% & \textbf{97.68\% $\pm$ 0.63\%} \\
    \midrule
    Avg & 42.34\% $\pm$ 0.70\%  & 55.07\% $\pm$ 0.24\% & 67.35\% $\pm$ 0.41\% & \textbf{68.57\% $\pm$ 0.37\%} \\
    \bottomrule
  \end{tabular}
\end{table}

From the general sight, Spark-Formalizer-X1-7B achieves an average pass rate of 68.57\%, attaining the SoTA performance across multiple datasets, outperforming the compared open-source models of the similar size. On the ProofNet-test set, Spark-Formalizer-X1-7B achieves a leading score of 78.85\%, a significant improvement of 3.04 percentage points over Goedel-Formalizer-V2-8B. Furthermore, it secures the best performance of 30.17\% on the challenging College CoT dataset. The comprehensive performance across benchmarks with varied academic levels and diverse sources demonstrates Spark-Formalizer-X1-7B's robust capability to formalize the mathematical propositions into formal language statements.

\subsection{Prover Evaluations}
\label{sec:prover_performance}

\paragraph{Main Results.} Evaluations among state-of-the-art formal
theorem-proving models are also conducted to verify the performance of formal proof generation. The experimental results under pass@32 are shown in Table~\ref{tab:prover-results}.

\begin{table}[H]\footnotesize
  \caption{Comparison results of provers on multiple test datasets under pass@32.}
  \label{tab:prover-results}
  \centering
  \begin{tabular}{lcccc}
    \toprule
    Benchmark   & \makecell{DeepSeek-Prover-\\V2-7B} & \makecell{Kimina-Prover-\\Distill-8B} & \makecell{Goedel-Prover\\-V2-8B} & \makecell{Spark-Prover\\-X1-7B} \\
    \midrule
    miniF2F-test & 75.6\% & 78.3\% & \textbf{84.6\%} & 75.0\% \\
    ProofNet-test & 23.0\% & 11.0\% & 19.4\% & \textbf{23.1\%} \\
    PutnamBench & 1.4\% & 2.9\% & 3.8\% & \textbf{4.7\%} \\
    CombiBench & 16.0\% & 6.0\% & 12.0\% & \textbf{24.0\%} \\
    FormalMATH-Lite & 51.8\% & 55.5\% & 55.3\% & \textbf{59.8\%} \\
    ProverBench & 49.0\% & 38.8\% & \textbf{52.0\%} & 47.4\% \\
    MathOlympiadBench & 8.9\% & 8.1\% & 10.8\% & \textbf{11.1\%} \\
    ExamFormal-Bench & 49.0\% & 45.3\% & 48.8\% & \textbf{51.2\%} \\
    \midrule
    Avg & 34.3\% & 30.7\% & 35.8\% & \textbf{37.0\%} \\
    \bottomrule
  \end{tabular}
\end{table}

The evaluations of the provers involve 8 test datasets for a comprehensive result. As illustrated in Table~\ref{tab:prover-results}, the proposed Spark-Prover-X1-7B achieves the highest score on average, demonstrating the advantages of Spark-Prover-X1-7B in formal theorem proving. Among the 8 benckmarks, Spark-Prover-X1-7B obtains the best performance in 6 datasets, which significantly indicates the general enhancement of Spark-Prover-X1-7B in formal theorem proving.

\paragraph{Results on Challenging Benchmarks.} 
For the challenging ProofNet-test, PutnamBench, and MathOlympiadBench, which are the top 3 difficult to be solved, Spark-Prover-X1-7B solves 23.1\%, 4.7\%, and 11.1\% problems respectively, under pass@32. These results illustrate that Spark-Prover-X1-7B obtains the best performance in ProofNet-test, PutnamBench, and MathOlympiadBench, surpassing other models in solving these challenging problems. The experimental results indicate that Spark-Prover-X1-7B exhibits the most comprehensive ability in proving the challenging theorems compared with the other provers.

\paragraph{Results on Combinatorial Problems.} 
CombiBench is a benckmark consisting of 100 combinatorial competition problems formalized from natutal languagr to the formal language statement~\cite{deepseekproverv2}. The evaluation result on CombiBench shows that Spark-Prover-X1-7B achieves a significant improvement in solving combinatorial mathematics, a known challenge for provers, solving 24\% of the problems on CombiBench, which is substantially higher than 16\% of DeepSeek-Prover-V2-7B.

\paragraph{Results on FormalMATH-Lite.} 
The FormalMATH-Lite dataset is a benchmark for formal theorem proving, covering diverse difficulty levels (from high school to undergraduate) and multiple mathematical categories (such as algebra, number theory, discrete mathematics, etc.). Spark-Prover-X1-7B solves 59.8\% of problems on FormalMATH-Lite under pass@32 and outperforms the compared provers. 

\paragraph{Results on ExamFormal-Bench.} We collect problems from actual exam scenarios and formalize them into formal language statements to construct the ExamFormal-Bench testdata. Spark-Prover-X1-7B also achieves the best performance of formal theorem proving on ExamFormal-Bench, which indicates the excellent ability in solving problems from math examinations.

\begin{table}[H]\footnotesize
  \caption{Comparison results of provers on PutnamBench.}
  \label{tab:putnam-results}
  \centering
  \begin{tabular}{llccc}
    \toprule
    \# & Models   & \makecell{Number-solved \up} & \makecell{Budget \down}  \\
    \midrule
    1 & \textbf{Spark-Prover-X1-7B} & 55 & pass@512 \\
    2 & DeepSeek-Prover-V2-671B & 47 & pass@1024 \\
    3 & \textbf{Spark-Prover-X1-7B} & 40 & pass@256 \\
    4 & \textbf{Spark-Prover-X1-7B} & 27 & pass@32  \\
    5 & Goedel-Prover-V2-8B & 25 & pass@32  \\
    6 & DeepSeek-Prover-V2-671B & 22 & pass@32  \\
    7 & Kimina-Prover-Distill-8B & 10 & pass@192 \\

    \bottomrule
  \end{tabular}
\end{table}

\paragraph{Results on PutnamBench.} Table~\ref{tab:putnam-results} summarizes the comparison results of different provers with various compute cost on the test data Putnam. Spark-Prover-X1-7B solves 27 problems, which is the most number compared with other models under pass@32. Meanwhile, Spark-Prover-X1-7B solves 55 problems under pass@512, which even surpasses the results of DeepSeek-Prover-V2-671B aunderpass@1024 by more 8 problems. These results indicate that Spark-Prover-X1-7B significantly outperforms the SoTA provers and shows comprehensive advantages of Spark-Prover-X1-7B in solving problems of PutnamBench.

\paragraph{Output length distribution.} 

\begin{figure}
  \centering
  \includegraphics[width=\textwidth]{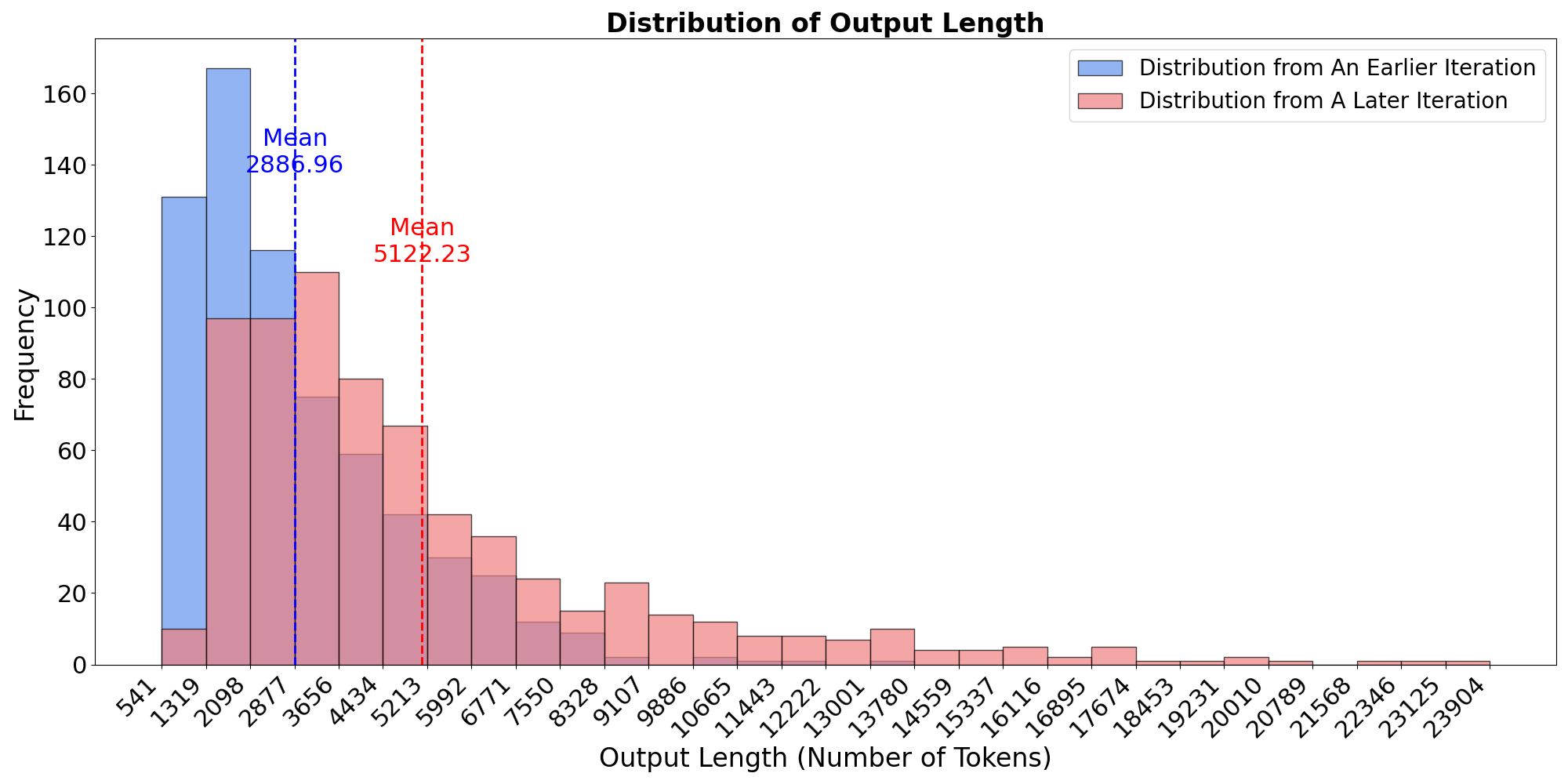}
  \caption{Evolution of output length distribution.}
  \label{fig:proof-length}
\end{figure}

As noted in prior work such as BFS-Prover\cite{bfsv1}, output length can, to some extent, reflect a prover's theorem proving capability. Therefore, we plot the combined distribution of output lengths (number of tokens) generated by Spark-Prover-X1-7B across all test sets, as shown in Figure~\ref{fig:proof-length}. The blue distribution represents output lengths from an earlier iteration model, while the red distribution represents those from a later, more refined model. As the expert iteration progresses, there exists a clear trend that the average output length of Spark-Prover-X1-7B shifts across the spectrum towards longer outputs, with the average length increasing from 2886.96 to 5122.23 tokens. This indicates that the model's reasoning capabilities are enhancing, allowing it to conduct longer and more effective chains of thought to solve complex problems.

Notably, even as the average output length increases, the model's output distribution remains predominantly composed of short-to-medium-length outputs. This indicates that our model retains its ability to generate concise proofs when appropriate. However, a common failure mode in reinforcement learning is that models can collapse to a narrow distribution of high-reward actions, thereby losing diversity in their reasoning patterns\cite{reinforcement}. We attribute this preservation of capability to our methodological decision to separate the final RL phase from the expert iteration loop. We only apply a single round of RL after a stable, multi-round SFT model has been obtained, rather than using RL to drive the iterative process itself.

\section{Conclusion}

In this paper, we present Spark-Prover-X1-7B with a three-stage training strategy that progressively refines our model, achieving new state-of-the-art results in multiple benchmarks on average. Our methodology integrates a robust continuous pre-training phase and a Supervised Fine-tuning (SFT) phase, both of which are encompassed within an expert iteration loop. This is followed by a final reinforcement learning stage to further enhance the \texttt{Spark-Prover-X1-7B}. This work is rooted in building a powerful foundational model, which is achieved by systematically expanding the training data and introducing a suite of diverse data tasks designed to enhance formal reasoning capabilities.

\bibliographystyle{plain}
\bibliography{ref}

\end{document}